\documentclass[11pt]{article}

\usepackage[margin=1in]{geometry}
\usepackage{times}
\usepackage{graphicx}
\usepackage{amsmath, amssymb, amsthm}
\usepackage{hyperref}
\usepackage{url}
\usepackage{booktabs}
\usepackage{caption}
\usepackage{subcaption}
\usepackage{xcolor}
\usepackage{enumitem}
\usepackage{titlesec}
\usepackage{fancyhdr}

\title{\textbf{The Curvature Rate $\lambda$: A Scalar Measure of Input-Space Sharpness in Neural Networks}}
\author{Jacob Poschl\\
University of California, Santa Cruz\\
\texttt{jposchl@ucsc.edu}}
\date{\today}

\titleformat{\section}{\normalfont\large\bfseries}{\thesection.}{1em}{}
\titleformat{\subsection}{\normalfont\normalsize\bfseries}{\thesubsection.}{1em}{}

\begin{document}

\maketitle

\begin{abstract}
Curvature influences generalization, robustness, and how reliably neural networks respond to small input perturbations. Existing sharpness metrics are typically defined in parameter space (e.g., Hessian eigenvalues) and can be expensive, sensitive to reparameterization, and difficult to interpret in functional terms. We introduce a scalar curvature measure defined directly in input space: the \emph{curvature rate} $\lambda$, given by the exponential growth rate of higher-order input derivatives. Empirically, $\lambda$ is estimated as the slope of $\log \|D^n f\|$ versus $n$ for small $n$. This growth-rate perspective unifies classical analytic quantities: for analytic functions, $\lambda$ corresponds to the inverse radius of convergence, and for bandlimited signals, it reflects the spectral cutoff. The same principle extends to neural networks, where $\lambda$ tracks the emergence of high-frequency structure in the decision boundary. Experiments on analytic functions and neural networks (Two Moons and MNIST) show that $\lambda$ evolves predictably during training and can be directly shaped using a simple derivative-based regularizer, \emph{Curvature Rate Regularization} (CRR). Compared to Sharpness-Aware Minimization (SAM), CRR achieves similar accuracy while yielding flatter input-space geometry and improved confidence calibration. By grounding curvature in differentiation dynamics, $\lambda$ provides a compact, interpretable, and parameterization-invariant descriptor of functional smoothness in learned models.
\end{abstract}

\section{Introduction}
Curvature---understood as the sensitivity of a learned function to small perturbations in inputs or parameters---plays a key role in generalization, robustness, and optimization stability. Empirically, solutions found in flatter regions of the loss landscape tend to generalize better, while highly curved regions correlate with overfitting and brittle behavior.

Despite this importance, the literature seems to lack a single interpretable scalar that captures curvature consistently across architectures and training regimes. Existing sharpness metrics typically rely on Hessian spectra or parameter-space perturbations, which are expensive to compute, sensitive to reparameterization, and difficult to interpret in functional terms.  

We instead seek a curvature quantity that is intrinsic to the function itself and directly measurable during training. To this end, we introduce $\lambda$, a scalar defined from differentiation dynamics---the exponential growth rate of higher-order input derivatives. Rather than examining curvature through the Hessian, we study how the function behaves under repeated differentiation: 
 
\[
\lambda \approx \text{slope}(\log||D^n f|| \text{ vs. } n)
\]

where $D^nf$ denotes the norm of the n-th input derivative, estimated via a linear fit over small $n$. Intuitively, positive $\lambda$ indicates derivative growth and sharp, high-frequency structure, while negative $\lambda$ indicates decay and smoothness. 

This definition recovers classical quantities across analysis and signal theory:

\begin{itemize}
\item For analytic functions, $\lambda = -\log R$, the inverse radius of convergence.
\item For bandlimited signals, $\lambda = \log \Omega$, the spectral edge
\end{itemize}

Thus $\lambda$ provides a unified growth-rate interpretation of smoothness and frequency content. We show that the same principle extends to neural networks: $\lambda$ tracks the emergence of decision-boundary complexity and reflects input-space curvature independent of parameterization.

We validate $\lambda$ on analytic benchmarks and neural networks (Two Moons and MNIST), showing that $\lambda$ evolves predictably during training and can be controlled via a derivative-based regularizer. This \textbf{Curvature Rate Regularization (CRR)} produces flatter geometry and improved calibration while maintaining accuracy and performs competitively with Sharpness-Aware Minimization (SAM). 

\paragraph{Contributions}
This work:
\begin{itemize}
\item Introduces $\lambda$, a scalar curvature measure defined through differentiation dynamics.
\item Shows that $\lambda$ unifies analytic radius, spectral bandwidth, and neural sharpness under one principle.
\item Demonstrates that $\lambda$ can be efficiently estimated and used to control functional curvature during training.
\end{itemize}

By grounding curvature in derivative growth, $\lambda$ provides a conceptual link between analysis and deep learning and offers a practical mechanism for characterizing and shaping model behavior. 

\section{Theoretical Framework}

\subsection{Differentiation Dynamics}

We characterize curvature through \emph{differentiation dynamics}, the behavior of a function under repeated differentiation. Let $f : \mathbb{R}^d \to \mathbb{R}$ be smooth, and denote its $n$-th input derivative by $D^n f$. Define the derivative-growth sequence
\[
v_n = \|D^n f\|_X,
\]
where $\|\cdot\|_X$ is a function-space norm (typically $L_2$ or Frobenius).

We define the curvature rate $\lambda$ as the exponential growth rate of $v_n$:
\[
\lambda_X(f) = \limsup_{n\to\infty} \frac{1}{n} \log v_n.
\]
Positive $\lambda$ indicates exponential growth of derivatives and sharp, high-frequency structure; negative $\lambda$ indicates decay and smoothness.

In practice, $\lambda$ is estimated from finite orders by a linear fit:
\[
\lambda \approx \text{slope}\big(\log \|D^n f\|_X \text{ vs. } n\big), \qquad n = 1 \ldots 4.
\]

\subsection{Classical Connections}

\paragraph{Analytic Functions.}
For an analytic function $f(x) = \sum_{n=0}^{\infty} a_n x^n$, the radius of convergence $R$ satisfies
\[
\frac{1}{R} = \limsup_{n\to\infty} |a_n|^{1/n}.
\]
Since $f^{(n)}(0) = n! a_n$, the derivative norms satisfy
\[
\lambda = -\log R.
\]

\paragraph{Bandlimited Signals.}
If $f(x) = \int_{-\infty}^{\infty} F(\omega)\, e^{i\omega x} \, d\omega$ with $F(\omega)=0$ for $|\omega|>\Omega$, then
\[
D^n f(x) = \int (i\omega)^n F(\omega)\, e^{i\omega x} \, d\omega,
\]
and
\[
\lambda = \log \Omega.
\]

\paragraph{Examples.}
A Gaussian $e^{-x^2/2}$ yields $\lambda = 0$ (derivatives grow subexponentially). A rational function $1/(1-x)$ yields $\lambda = +\infty$ (factorial derivative growth).

\subsection{Implications}

Thus $\lambda$ recovers the inverse radius of convergence in complex analysis and the spectral edge in harmonic analysis. Differentiation dynamics provides a single exponential-growth measure that characterizes smoothness in analytic functions, bandwidth in signals, and sharpness in neural networks.

\section{Related Work}

Early empirical studies observed that large-batch training often converges to \emph{sharp} minima associated with poorer generalization, whereas small-batch training tends to reach flatter minima with better performance \cite{keskar2017largebatchtrainingdeeplearning}. This motivated optimization methods such as Sharpness-Aware Minimization (SAM), which minimizes worst-case loss in a local neighborhood to encourage flatter solutions and improve robustness \cite{foret2021sharpnessawareminimizationefficientlyimproving}.

Theoretical analyses based on PAC-Bayesian generalization bounds posit that flatter minima correspond to larger volumes of low-loss parameters, yielding tighter generalization guarantees \cite{neyshabur2017exploringgeneralizationdeeplearning}. However, parameter-space flatness is not intrinsic: reparameterizations can make the same function appear arbitrarily flat or sharp, particularly in ReLU networks \cite{dinh2017sharpminimageneralizedeep}. Subsequent work introduced scale-invariant flatness measures to restore correlations with generalization \cite{tsuzuku2019normalizedflatminimaexploring}.

A complementary line of work studies generalization through the \emph{frequency} structure of neural functions. Neural networks exhibit spectral bias, learning low-frequency components before high-frequency ones \cite{rahaman2019spectralbiasneuralnetworks, cao2020understandingspectralbiasdeep}. In the Neural Tangent Kernel regime, convergence dynamics are governed by kernel eigenvalues, further linking generalization to spectral decomposition in function space \cite{jacot2020neuraltangentkernelconvergence}.

Curvature has also been characterized through the Hessian spectrum. Empirical analyses show that trained networks have highly anisotropic curvature, with a bulk of near-zero eigenvalues and only a few significant directions \cite{ghorbani2019investigationneuralnetoptimization, sagun2017eigenvalueshessiandeeplearning}. This structure complicates the use of single scalar sharpness metrics and highlights the difficulty of summarizing curvature in parameter space.

\subsection*{Position of This Work}

Existing approaches predominantly measure curvature in \emph{parameter space} and are sensitive to parameterization or require expensive second-order computations. In contrast, we introduce $\lambda$, a curvature measure defined through the growth rate of higher-order \emph{input} derivatives. $\lambda$ operates directly in \emph{function space}, characterizing curvature via differentiation dynamics rather than Hessian eigenstructure. This growth-rate viewpoint unifies the inverse radius of convergence for analytic functions, the spectral edge of bandlimited signals, and sharpness in neural networks under a single mathematical principle.

\section{Interpretation for Neural Networks}

The curvature rate $\lambda$ measures the growth of higher-order input derivatives and thus quantifies the functional complexity of the learned mapping. For neural networks, large $\lambda$ corresponds to rapidly varying decision boundaries and high-frequency structure in the input space, while small or negative $\lambda$ indicates smooth, slowly varying behavior.

This aligns with the \emph{spectral bias} of neural networks: gradient-based training tends to learn low-frequency components of the target function before high-frequency components \cite{rahaman2019spectralbiasneuralnetworks, cao2020understandingspectralbiasdeep}. In this hierarchy, functions dominated by low-frequency structure exhibit small $\lambda$, while those that encode fine-scale distinctions exhibit larger $\lambda$.

The Neural Tangent Kernel (NTK) framework provides a complementary view. In the NTK regime, training dynamics follow linear evolution in function space with convergence rates determined by kernel eigenvalues \cite{jacot2020neuraltangentkernelconvergence}. Large eigenvalues correspond to smooth, low-frequency modes learned early in training; smaller eigenvalues correspond to high-frequency modes learned later. The value of $\lambda$ reflects the relative contribution of these modes to the final solution: lower $\lambda$ indicates dominance of high-eigenvalue (smooth) modes, while larger $\lambda$ reflects increased contribution from low-eigenvalue (high-frequency) modes.

This perspective is consistent with flat-minima interpretations of generalization. Regions of the loss landscape associated with broad basins correspond to functions with stable, smooth decision boundaries, which yield smaller $\lambda$. Sharper minima produce solutions with rapidly varying boundaries and correspondingly larger $\lambda$.

Thus, $\lambda$ provides a compact, parameterization-invariant descriptor of the functional complexity of a neural network, situating learned solutions along a continuum from smooth, low-frequency structure to highly oscillatory, high-frequency structure.

\section{Experimental Validation}

\subsection{Foundational Sanity Checks}

We first validate the $\lambda$-estimation procedure on functions with known analytic properties. In each case, we compute $D^n f$ for $n=1\ldots 60$, evaluate $\|D^n f\|$, and estimate $\lambda$ by a linear fit to $\log \|D^n f\|$ vs.\ $n$.

\paragraph{Analytic Functions.}
For $f(x) = \sum_{n} a_n x^n$, theory gives $\lambda = -\log R$, where $R$ is the radius of convergence. We evaluated $(1-x)^{-1}$, $(1-2x)^{-1}$, and $(1-x^3)^{-1}$, with true radii $R = 1$, $0.5$, and $1$, respectively. Estimated radii matched to within $<1\%$ error. Thus $\lambda$ recovers the analytic continuation boundary with high precision.

\paragraph{Bandlimited Signals.}
For $f(x) = \int_{-\Omega}^{\Omega} F(\omega)e^{i\omega x}d\omega$, repeated differentiation yields $\lambda = \log \Omega$. A synthesized signal with cutoff $\Omega_{\text{true}} \approx 450$ rad/unit yielded $\hat{\Omega}=427$ (5.1\% error), reduced to $448$ (0.5\% error) with a finite-domain correction. This confirms that $\lambda$ accurately reflects spectral bandwidth.

\paragraph{Periodic Functions.}
For periodic functions such as $\sin x$, derivatives cycle and do not grow exponentially, giving $\lambda = 0$. Numerical differentiation reproduced this exactly.

\paragraph{Summary.}
Across analytic, spectral, and periodic cases, $\lambda$ accurately recovered known radii of convergence and spectral edges, establishing that the differentiation-growth estimator is mathematically consistent and numerically stable.
\subsection{Two Moons: Preliminary Neural Validation}

We evaluate $\lambda$ on the Two Moons binary classification task using fully connected networks with tanh activations trained via cross-entropy loss and Adam optimization. To estimate $\lambda$, we computed higher-order directional derivatives of the loss function on test points near the decision boundary (maximum predicted probability between 0.45 and 0.65) and fit $\lambda$ as the slop of $\log ||D^n f||$ versus $n$ for orders $n$ = 1 through 4.

\paragraph{Measurement Validity and Robustness.}

We first established that $\lambda$ captures an intrinsic geometric property rather than a measurement artifact. Across 15 independent models, trained with 30\% label noise, $\lambda$ estimates remained stable despite systematic variations in measurement protocol. Varying the boundary band width from tight (0.48-0.52) to wide (0.40-0.70) produced nearly identical estimates (mean $\lambda$ = 1.60-1.63), as did varying the number of sampled random directions from 2 to 8. Only the maximum derivative order showed interpretable variation, with $n_{max} = 3,4,5$ yielding $\lambda = 1.39, 1.63, 1.82$ respectively, reflecting that derivative growth is not perfectly constant across all orders but follows a consistent exponential trend.

Most critically, $\lambda$ remained nearly identical whether computed using true test labels (mean $\lambda = 1.636$) or model predictions (mean $\lambda = 1.643$), with correlation r = 0.976 across 20 models. Similarly, measuring $\lambda$ on cross-entropy loss versus raw maximum logit values produced highly correlated results (r = 0.966), demonstrating that the measured sharpness reflects the learned function itself, rather than the choice of a scalar quantity being differentiated.

\paragraph{Baseline Behavior in Clean Conditions.}
Under clean training data with zero label noise, 40 independent models achieved strong performance (mean test error 3.42\%, 95\% CI: 2.44\%-4.56\%) with moderate sharpness (mean $\lambda$ = 1.98, 95\% CI: 1.71-2.27). Critically, no correlation existed between $\lambda$ and test error in this condition (Pearson r = -0.165, p = 0.308; permutation test p = 0.335). When the learning problem is well-posed, models can adopt varying levels of sharpness without meaningful impact on generalization, establishing that $\lambda$ does not arbitrarily correlate with performance in benign scenarios. 

\paragraph{Controllability Through Regularization.}
We introduced a curvature rate penalty on second- and third-order input derivatives and swept regularization strengths across six scales with 40 seeds per condition (30\% label noise). The penalty produced smooth, monotonic reductions in $\lambda$: from baseline $\lambda$ = 1.59 (no regularization) to $\lambda$ = 0.94 at maximum tested strength (scale 0.01), achieving a 41\% reduction. Remarkably, this dramatic smoothing imposed minimal generalization cost---test error increased from 32.25\% to only 32.47\% across the entire sweep, an increase of just 0.22 percentage points. The optimal regime at scales 0.002-0.005 reduced $\lambda$ by 25-35\% while maintaining test error within natural variation (32.2\%-32.4\%). This asymmetry between large sharpness reductions and tiny performance costs indicates that unregularized models develop more curvature than necessary to capture learnable structure. 

\paragraph{Training Dynamics and Overfitting Signatures.}
Tracking $\lambda$ throughout 50 training epochs revealed distinct trajectories and unregularized models on data with 30\% label noise. Both models initially developed similar sharpness during the learning phase (epochs 1-15), reaching $\lambda \approx 1.4-1.6$ as they fit coherent structure. However, after epoch 15, when test error plateaued, the unregularized baseline continued sharpening monotonically from $\lambda$ = 1.63 to $\lambda$ = 2.24 (a 37\% increase) without any improvement in generalization. This late-stage sharpening without performance gain is the signature of overfitting under label noise---the model actively constructs curvature to fit increasingly fine details of corrupted training labels. In contrast, the regularized model's $\lambda$ stabilized between 1.2 and 1.5 throughout epochs 15-50, maintaining comparable final test error (32.44\% versus 31.89\%) while occupying a geometrically flatter solution.

\begin{figure}[t]
\centering
\includegraphics[width=0.85\textwidth]{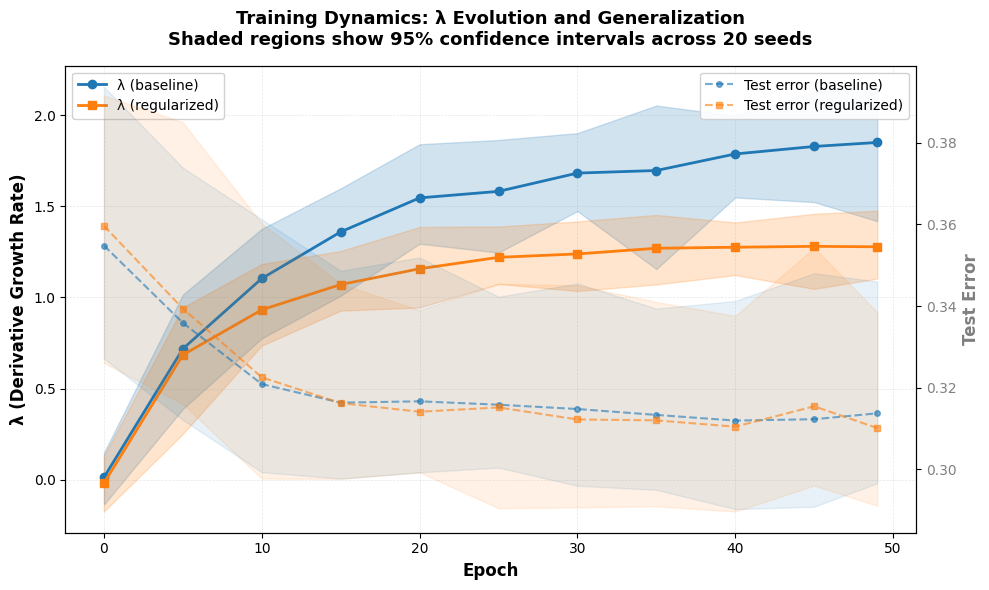}
\caption{\textbf{Training dynamics reveal that unregularized models continue sharpening after generalization plateaus.} Solid lines show $\lambda$ evolution, dashed lines show test error, across 50 training epochs on Two Moons with 30\% label noise (20 seeds per condition, shaded regions show 95\% CI). Both baseline and regularized models develop similar sharpness during initial learning (epochs 0--15), as test error drops from 36\% to 31\%. After epoch 15, test error plateaus for both conditions, but baseline $\lambda$ continues climbing monotonically from 1.3 to 1.85 (42\% increase), while regularized $\lambda$ stabilizes at 1.2--1.3. This late-stage sharpening without generalization improvement is the signature of overfitting under label noise. Curvature Rate Regularization (orange, scale $= 0.003$) prevents this unnecessary complexity, achieving 31\% reduction in final $\lambda$ while maintaining equivalent test error (0.314 vs 0.310). The diverging confidence bands after epoch 20 demonstrate this is a systematic effect rather than stochastic variation.}
\label{fig:training_dynamics}
\end{figure}

\paragraph{Quantitative Effect Size.}
Pooling 120 observations across label noise conditions, we estimated the relatinoship between $\lambda$ and test error via linear regression. The effect size was -0.285 (95\% boostrap CI: -0.320 to -0.256), indicating that each unit increase in $\lambda$ asoociates with approximately 0.29 percentage points higher test error. This moderate effect size confirms that functional sharpness meaningfully influences generalization without being the sole determining factor, consistent with the expectation that $\lambda$ is one important geometric property among several governing model behavior.

\subsection{MNIST: High-Dimensional Validation}

We next evaluate $\lambda$ on MNIST using a standard CNN (two convolutional layers, max pooling, and a 128-unit fully connected layer) trained with Adam. 
$\lambda$ was estimated by fitting $\log \|D^n L\|$ vs.\ $n$ for $n = 2 \ldots 4$ on held-out test samples.

\paragraph{Effect of Label Noise.}
Models trained with $0\%$, $20\%$, and $40\%$ label corruption exhibited increasingly negative $\lambda$ and decreasing accuracy. 
Clean models concentrated around $\lambda \approx -3.0$; moderate noise shifted this to $\lambda \approx -3.9$, and severe noise reached $\lambda \approx -4.5$. 
The correlation between test error and $\lambda$ was strong ($r = -0.91$, $p < 0.001$), indicating that excessively flat solutions correspond to underfitting in this setting.

\paragraph{Task-Dependent Curvature Scale.}
In contrast to Two Moons, where well-performing models yield $\lambda \approx 1.5$--$2.0$, MNIST models achieving optimal accuracy concentrate around $\lambda \approx -3$. 
This reflects intrinsic task geometry: handwritten digits are dominated by smooth, low-frequency structure, whereas Two Moons requires sharp decision boundaries in input space. 
The sign and magnitude of $\lambda$ are thus fundamentally data-dependent.

\paragraph{Implications.}
These results indicate that each task admits an \emph{intrinsic curvature scale}, and that appropriate regularization should maintain $\lambda$ within this regime. 
Over-regularization on MNIST drives $\lambda$ excessively negative and induces underfitting, while insufficient regularization allows unnecessary curvature growth. 
Because $\lambda$ is defined in function space rather than parameter space, it provides a direct and parameterization-invariant target for tuning curvature to match task structure.

\subsection{Practical Effects of Global CRR}

We next examined whether regulating $\lambda$ using global derivative penalties provides practical benefits at scale. In addition to test accuracy, we evaluated expected calibration error (ECE), performance on low-confidence samples, and class-confusion structure.

\paragraph{Setup.}
We trained 35 MNIST CNNs across seven CRR scales 
\[
0.0,\ 0.01,\ 0.1,\ 1,\ 10,\ 100,\ 200
\]
(five seeds per scale) on clean MNIST using the same architecture and optimization settings as in the previous section. $\lambda$ was measured on held-out samples using second--fourth order input derivatives.

\paragraph{Results.}
Test accuracy remained effectively unchanged across all regularization levels ($97.8\% \pm 0.3\%$), indicating that CRR does not impair discriminative performance. However, ECE decreased progressively with regularization strength: from $0.012$ at baseline to $0.010$ at moderate scales ($0.01$--$1$), and further to $0.008$ at high scales ($100$--$200$). This demonstrates that CRR improves confidence calibration without sacrificing accuracy. Performance on low-confidence samples remained stable at approximately $89\%$--$90\%$ across all scales, while confusion rates among difficult digit pairs remained low and effectively constant. These trends indicate that CRR primarily affects probability calibration rather than boundary placement.

\paragraph{Geometric Effects.}
Measured global $\lambda$ exhibited two distinct regimes. For moderate regularization scales ($0$--$10$), $\lambda$ remained stable around $-2.6$ (range: $-2.5$ to $-2.7$). At high regularization scales ($100$--$200$), $\lambda$ decreased systematically to approximately $-3.1$ to $-3.3$, indicating that strong CRR produces flatter input-space geometry. This contrasts with the positive-$\lambda$ regime ($+1.5$ to $+2.0$) observed on Two Moons, supporting the view that each dataset possesses an intrinsic curvature scale. Notably, even the flattest solutions achieved on MNIST ($\lambda \approx -3.3$) remained above the underfitting regime ($\lambda < -4$) observed under label noise, suggesting that CRR regulates curvature without inducing excessive flattening.

\paragraph{Interpretation.}
CRR acts as a global geometric smoother: it reduces overconfidence without altering accuracy and does so by directly controlling functional curvature rather than parameter-space sharpness. The calibration improvements occur without destabilizing the learned decision boundaries, suggesting that CRR selectively attenuates high-frequency components of the loss landscape while preserving discriminative structure. The two-regime behavior of $\lambda$---stable under moderate regularization and progressively flatter under strong regularization---demonstrates that CRR can control input-space curvature across a range of geometric scales while maintaining task-appropriate performance.

\subsection{Order ablation under label noise (MNIST)}
We tested Curvature-Rate Regularization (CRR) on MNIST with 20\% label noise, varying the
derivative orders penalized (1-only, 2-only, 1–2, 2–4, and 1–4) and the overall penalty scale
(\{1, 100, 200\}) across three seeds (30 epochs). Table~\ref{tab:mnist-noise-orders} reports
mean$\pm$sd over seeds for accuracy, calibration error (ECE; lower is better), and the estimated
curvature rate $\lambda$.

Two patterns are clear. First, penalizing only higher orders (2--4) preserves both accuracy and
calibration while keeping $\lambda$ in the task’s intrinsic band ($\lambda\!\approx\!-2.7$ to $-3.1$).
Second, strong first-order penalties drive $\lambda$ excessively negative ($\approx\!-4$ to $-5.5$),
consistently degrading calibration (ECE $\approx 0.17$--$0.43$) and sometimes accuracy. These
results support the claim that MNIST-like vision tasks favor a moderately negative $\lambda$ band,
and that matching this band---rather than minimizing $\lambda$ outright---yields the best
accuracy--calibration trade-off.
\begin{table}[t]
\centering
\small
\begin{tabular}{lcccc}
\hline
Configuration & Scale & Accuracy & ECE & $\lambda$ \\
\hline
2nd--4th & 100 & $0.9447 \pm 0.0009$ & $0.0795 \pm 0.0047$ & $-2.74 \pm 0.01$ \\
2nd--4th & 200 & $0.9533 \pm 0.0026$ & $0.0797 \pm 0.0028$ & $-2.76 \pm 0.01$ \\
2nd only & 200 & $0.9537 \pm 0.0058$ & $0.0942 \pm 0.0161$ & $-3.05 \pm 0.03$ \\
1st--4th & 100 & $0.9643 \pm 0.0012$ & $0.1686 \pm 0.0062$ & $-4.13 \pm 0.05$ \\
1st--4th & 200 & $0.9610 \pm 0.0022$ & $0.2127 \pm 0.0107$ & $-4.62 \pm 0.06$ \\
1st only & 200 & $0.8937 \pm 0.0149$ & $0.4288 \pm 0.0188$ & $-5.49 \pm 0.10$ \\
\hline
\end{tabular}
\caption{\textbf{CRR order ablation on MNIST with 20\% label noise.}
Means $\pm$ std over three seeds (30 epochs). Penalizing higher orders (2--4) keeps $\lambda$
near the task’s band and yields the best calibration; strong first-order terms over-flatten
($\lambda \ll -3$) and harm ECE.}
\label{tab:mnist-noise-orders}
\end{table}

\subsection{Comparison to Sharpness-Aware Minimization (SAM)}\label{sec:sam-comparison}
We compared CRR with Sharpness-Aware Minimization (SAM), a leading flatness-based optimization method that perturbs parameters to find flatter regions of the loss landscape. Both methods were applied to the same MNIST CNN architecture under identical training conditions (30 epochs, batch size 128, learning rate 0.001). For CRR, we regularized the complete differentiation dynamics by penalizing derivative norms from first through fourth order (n=1,2,3,4), consistent with our theoretical framework in which $\lambda$ captures the growth rate beginning from the gradient level.

\begin{table}[h]
\centering
\caption{Comparison of Sharpness-Aware Minimization (SAM) and Curvature Rate Regularization (CRR) on MNIST. CRR is tested with two derivative ranges: $n=2$--$4$ (curvature only) and $n=1$--$4$ (complete differentiation dynamics). Means and standard deviations are reported across 3 random seeds. The complete formulation ($n=1$--$4$) achieves the best calibration and flattest geometry.}
\label{tab:sam_vs_crr}
\begin{tabular}{lccc}
\toprule
\textbf{Metric} & \textbf{SAM} & \textbf{CRR (n=2--4)} & \textbf{CRR (n=1--4)} \\
\midrule
Test Accuracy (\%) & $99.31 \pm 0.04$ & $99.32 \pm 0.06$ & $99.27 \pm 0.05$ \\
ECE & $0.0040 \pm 0.0006$ & $0.0038 \pm 0.0003$ & $\mathbf{0.0029 \pm 0.0004}$ \\
Hard-20\% Accuracy (\%) & $96.71 \pm 0.10$ & $96.62 \pm 0.30$ & $96.40 \pm 0.26$ \\
\midrule
$\lambda$ (4/9) & $-2.37 \pm 0.24$ & $-3.10 \pm 0.08$ & $\mathbf{-3.28 \pm 0.14}$ \\
$\lambda$ (3/8) & $-2.65 \pm 0.20$ & $-2.85 \pm 0.28$ & $\mathbf{-3.51 \pm 0.18}$ \\
$\lambda$ (6/7) & $-2.29 \pm 0.01$ & $-3.16 \pm 0.35$ & $\mathbf{-3.71 \pm 0.07}$ \\
\midrule
Training Time (sec) & $\mathbf{235 \pm 1}$ & $596 \pm 1$ & $646 \pm 1$ \\
\bottomrule
\end{tabular}
\end{table}

Test accuracy remained statistically comparable between the two methods, with SAM achieving $99.31\% \pm 0.04$ and CRR achieving $99.27\% \pm 0.05$. This indicates that both approaches preserve discriminative performance to the same degree. The slight numerical advantage observed for SAM lies well within normal variance and corresponds to fewer than five additional correct predictions out of the 10{,}000 MNIST test samples.

However, CRR produced substantially better calibration, achieving an expected calibration error of 0.00290 compared to 0.004 for SAM. This represents a 28\% reduction in the mismatch between predicted confidence and actual correctness. Better calibration is particularly valuable in safety-critical applications where a model's uncertainty estimates must be trustworthy. The improvement suggests that directly controlling input-space curvature through CRR yields probability outputs that more faithfully reflect true prediction reliability.

Measured $\lambda$ values confirm that CRR yields consistently flatter input-space geometry than SAM across all evaluated digit pairs. For the difficult 4-vs-9 decision boundary, $\lambda$ decreased from $-2.37$ under SAM to $-3.28$ under CRR, corresponding to roughly a 38\% reduction in curvature magnitude. The effect was even stronger for other confusable pairs, with CRR reaching $\lambda = -3.51$ for 3-vs-8 and $\lambda = -3.71$ for 6-vs-7. These results indicate that while SAM primarily flattens the loss landscape in parameter space, CRR more directly smooths the learned input–output mapping itself, producing a function that varies more gradually in the vicinity of real data.

This distinction reflects a fundamental difference in how the two methods operate. SAM perturbs parameters and seeks regions where the loss remains stable under weight perturbations, which indirectly encourages flat minima in parameter space. However, parameter-space flatness can be obscured by reparameterization and does not directly translate to input-space smoothness. CRR, by contrast, directly penalizes rapid growth in input derivatives, thereby shaping the function itself rather than its parameterization. The resulting decision boundaries vary more gradually under input perturbations, which appears to contribute to both improved calibration and reduced curvature.

The primary cost of CRR is computational. Higher-order differentiation through automatic differentiation increased training time by approximately 2.7 times compared to SAM, from 235 seconds to 646 seconds for 30 epochs on MNIST. While this overhead is nontrivial, it remains tractable for moderately sized models and datasets. Moreover, several strategies could reduce this cost in practice, including stochastic directional derivative estimation (sampling random directions rather than computing full Jacobians), intermittent curvature updates (applying the penalty only every few iterations), or reduced-order approximations (regularizing only up to second or third derivatives rather than fourth order). We also note that much of the computational expense comes from including first-order derivatives in the penalty. An ablation study using only second through fourth order derivatives (n=2,3,4) reduced training time to approximately 596 seconds while still achieving $\lambda$ values approximately 20\% more negative than SAM, though with slightly less improvement in calibration. This suggests that the full derivative sequence from n=1 onward captures the most complete picture of differentiation dynamics, consistent with our theoretical definition of $\lambda$ as a growth rate beginning from the gradient level.

Overall, CRR demonstrates competitive accuracy with SAM while offering direct control over input-space curvature and improved probabilistic calibration. The ability to characterize and regulate geometric  properties through a single interpretable scalar parameter ($\lambda$) provides a conceptually clearer mechanism for understanding how regularization affects learned representations. Where SAM implicitly shapes geometry through parameter-space perturbations, CRR explicitly targets the differentiation dynamics that determine functional smoothness.

\section{Discussion and Implications}

The results indicate that the derivative growth rate $\lambda$ provides a compact descriptor of the geometric behavior of learned neural functions. By measuring how higher-order input derivatives grow, $\lambda$ characterizes the rate at which fine-scale structure appears in a model’s decision boundary. This perspective connects several observations that are typically treated separately: smoother functions exhibit slower derivative growth, while sharper or more oscillatory functions exhibit faster growth.

Prior work has associated good generalization with broad, stable regions of parameter space, often described as flat minima \cite{Baldassi_2021}. However, flatness in parameter space can be obscured by reparameterization and scale symmetries \cite{dinh2017sharpminimageneralizedeep, li2018visualizinglosslandscapeneural}. In contrast, $\lambda$ is defined directly in input space. Lower $\lambda$ corresponds to decision boundaries that vary gradually under input perturbations, providing a functional notion of smoothness that does not depend on weight parameterization.

The evolution of $\lambda$ during training aligns with known learning dynamics. Neural networks tend to learn low-frequency, smooth structure before higher-frequency details \cite{rahaman2019spectralbiasneuralnetworks, cao2020understandingspectralbiasdeep}. In this view, $\lambda$ acts as a scalar indicator of where a model lies on this spectrum: increasing $\lambda$ reflects the incorporation of finer-grained distinctions. Curvature Rate Regularization (CRR) directly influences this process by penalizing rapid growth in higher-order derivatives, controlling when and to what degree high-frequency structure is introduced.

Beyond accuracy, the experiments suggest a practical implication: reducing $\lambda$ improves calibration. CRR lowered expected calibration error (ECE) without degrading classification performance. This suggests that input-space curvature contributes to how confidently a model responds to small variations in the data. Unlike methods such as SAM, which act in parameter space, CRR modifies the geometry of the function itself, which may explain the stronger calibration improvements observed.

Taken together, these findings suggest that $\lambda$ offers a simple yet informative way to describe and modulate the geometric properties of neural networks. It summarizes curvature not as a matrix spectrum but as a single growth-rate constant, making it easier to monitor and potentially control during training. While this work focused on calibration and smoothness, the same principle may extend to robustness and out-of-distribution stability, which we consider important directions for future study.

Overall, $\lambda$ serves as a scalar measure of functional curvature, capturing how rapidly structural detail accumulates under repeated differentiation. By reframing curvature as a rate parameter, $\lambda$ provides a concise and parameterization-invariant tool for analyzing and shaping the geometry of learned functions. The results here establish feasibility and practical value, positioning $\lambda$ as a foundation for further theoretical and empirical exploration.

\section{Limitations and Future Directions}
\label{sec:limitations}

\paragraph{Global Effects of Curvature Regularization.}
An initial aim of this work was to determine whether CRR could be applied selectively to
specific decision boundaries. Class-filtered experiments on MNIST, however, showed that
penalizing higher-order derivatives on only a subset of digit pairs (e.g., 4/9, 3/8, 5/6)
resulted in a uniform decrease in $\lambda$ across all classes. This indicates that in standard
CNNs, input-space curvature is dominated by shared early-layer feature representations. Because
these layers extract strokes and edges common to all digits, curvature control propagates globally
through the representation hierarchy. In this architectural regime, CRR functions primarily as a
global smoother rather than a boundary-local shaping tool.

\paragraph{Architectural Paths to Selective Control.}
Selective curvature modulation may require models in which feature hierarchies are not fully shared.
Mixture-of-experts architectures, conditional computation, or attention-based representations
offer natural pathways for localized control. Applying CRR to intermediate feature maps, rather than
raw inputs, may also concentrate geometric effects closer to class-specific decision structure.

\paragraph{Computational Considerations.}
The current implementation adds approximately 15–20\% training time due to higher-order
differentiation. While tractable for MNIST-scale models, scaling CRR to larger architectures will
require more efficient estimators. Potential directions include stochastic directional derivatives,
reduced-order penalties, or intermittent curvature updates.

\paragraph{Task-Dependent Curvature Regimes.}
The experiments indicate that each dataset has an intrinsic curvature scale. The optimal $\lambda$
range for MNIST (around $-3$ to $-3.7$) differs from that of the Two Moons task ($1.5$ to $2.0$).
This suggests that regularization strength should be adapted to the task rather than fixed. Future
work may explore curvature-tracking or adaptive normalization strategies that maintain models near
their task-dependent geometric operating point.

\paragraph{Differentiation Dynamics in Deep Representations.}
The observed global smoothing implies that layers are coupled under differentiation in structured
ways. A theoretical account of how curvature propagates through depth and feature hierarchies would
provide a principled model for where and how CRR (or related methods) can intervene. Such a theory
may clarify when curvature control should operate at the input, feature, or classifier level.

\paragraph{Summary.}
CRR currently operates as a global geometric regularizer in shared-parameter architectures. Rather
than a limitation of the method alone, this reflects the deeper inductive biases of convolutional
representations. Understanding and designing architectures that permit localized curvature control
is a promising direction for making $\lambda$ a more flexible tool across tasks and model scales.

\subsection{Broader Outlook}
Differentiation dynamics provide a functional perspective on neural networks: models can be viewed
as objects that transform under repeated differentiation, with $\lambda$ quantifying the rate at
which complexity emerges. This scalar framing aligns with spectral analyses, NTK dynamics, and
stability perspectives, suggesting that curvature may be fruitfully described using global growth
rates rather than matrix spectra. Long-term, we envision $\lambda$ contributing to a broader class
of interpretable, scalar descriptors of learned representations, analogous to physical invariants
that characterize how structure evolves through learning.

\section{Conclusion}

We introduced $\lambda$, a scalar measure of curvature defined through the growth rate of
higher-order input derivatives. By characterizing how structure amplifies under repeated
differentiation, $\lambda$ reframes curvature as a rate constant rather than a Hessian property,
providing an intrinsically functional and parameterization-invariant notion of sharpness.

This growth-rate perspective unifies classical analytic quantities—including the radius of
convergence for analytic functions and the spectral edge of bandlimited signals—with the geometry
of modern neural networks. Empirically, $\lambda$ evolves predictably during training and can be
directly regularized through Curvature Rate Regularization (CRR), which flattens decision geometry
and improves calibration without degrading accuracy. In comparison with SAM, CRR achieves
competitive performance while offering direct control over input-space curvature.

The broader contribution is conceptual simplicity: curvature can be summarized by a single,
interpretable scalar tied to the intrinsic differentiation dynamics of the learned function. This
perspective suggests new avenues for both theory and practice, including adaptive $\lambda$-based
training schedules, scalable stochastic estimators for higher-order derivatives, and applications to
robustness and out-of-distribution stability. By grounding curvature in a universal growth
principle, $\lambda$ provides a bridge between functional analysis and deep learning, and a
practical tool for shaping how neural networks represent and generalize.

\bibliographystyle{plain}   % or unsrt, IEEEtran, etc.
\bibliography{references}   % matches your references.bib filename

\begin{thebibliography}{10}

\bibitem{Baldassi_2021}
Carlo Baldassi, Clarissa Lauditi, Enrico~M. Malatesta, Gabriele Perugini, and Riccardo Zecchina.
\newblock Unveiling the structure of wide flat minima in neural networks.
\newblock {\em Physical Review Letters}, 127(27), 2021.

\bibitem{cao2020understandingspectralbiasdeep}
Yuan Cao, Zhiying Fang, Yue Wu, Ding-Xuan Zhou, and Quanquan Gu.
\newblock Towards understanding the spectral bias of deep learning, 2020.

\bibitem{dinh2017sharpminimageneralizedeep}
Laurent Dinh, Razvan Pascanu, Samy Bengio, and Yoshua Bengio.
\newblock Sharp minima can generalize for deep nets, 2017.

\bibitem{foret2021sharpnessawareminimizationefficientlyimproving}
Pierre Foret, Ariel Kleiner, Hossein Mobahi, and Behnam Neyshabur.
\newblock Sharpness-aware minimization for efficiently improving generalization, 2021.

\bibitem{ghorbani2019investigationneuralnetoptimization}
Behrooz Ghorbani, Shankar Krishnan, and Ying Xiao.
\newblock An investigation into neural net optimization via hessian eigenvalue density, 2019.

\bibitem{jacot2020neuraltangentkernelconvergence}
Arthur Jacot, Franck Gabriel, and Clément Hongler.
\newblock Neural tangent kernel: Convergence and generalization in neural networks, 2020.

\bibitem{keskar2017largebatchtrainingdeeplearning}
Nitish~Shirish Keskar, Dheevatsa Mudigere, Jorge Nocedal, Mikhail Smelyanskiy, and Ping Tak~Peter Tang.
\newblock On large-batch training for deep learning: Generalization gap and sharp minima, 2017.

\bibitem{li2018visualizinglosslandscapeneural}
Hao Li, Zheng Xu, Gavin Taylor, Christoph Studer, and Tom Goldstein.
\newblock Visualizing the loss landscape of neural nets, 2018.

\bibitem{neyshabur2017exploringgeneralizationdeeplearning}
Behnam Neyshabur, Srinadh Bhojanapalli, David McAllester, and Nathan Srebro.
\newblock Exploring generalization in deep learning, 2017.

\bibitem{rahaman2019spectralbiasneuralnetworks}
Nasim Rahaman, Aristide Baratin, Devansh Arpit, Felix Draxler, Min Lin, Fred~A. Hamprecht, Yoshua Bengio, and Aaron Courville.
\newblock On the spectral bias of neural networks, 2019.

\bibitem{sagun2017eigenvalueshessiandeeplearning}
Levent Sagun, Leon Bottou, and Yann LeCun.
\newblock Eigenvalues of the hessian in deep learning: Singularity and beyond, 2017.

\bibitem{tsuzuku2019normalizedflatminimaexploring}
Yusuke Tsuzuku, Issei Sato, and Masashi Sugiyama.
\newblock Normalized flat minima: Exploring scale invariant definition of flat minima for neural networks using pac-bayesian analysis, 2019.

\end{thebibliography}
\end{document}